\documentclass[10pt,twocolumn,letterpaper]{article}

\usepackage{iccv}
\usepackage{times}
\usepackage{epsfig}
\usepackage{graphicx}
\usepackage{amsmath}
\usepackage{amssymb}

\usepackage[pagebackref=true,breaklinks=true,letterpaper=true,colorlinks,bookmarks=false]{hyperref}

\usepackage{booktabs}

\usepackage{algorithmic}
\usepackage{algorithm}

\usepackage{xcolor}
\usepackage{multirow}
\usepackage{tablefootnote}
\usepackage{colortbl}
\usepackage{chngcntr}
\usepackage{inconsolata}
\newcounter{magicrownumbers}
\newcommand\rownumber{\stepcounter{magicrownumbers}\arabic{magicrownumbers}}

\usepackage{pifont}\newcommand{\cmark}{\ding{51}}\newcommand{\xmark}{\ding{55}}\definecolor{LightCyan}{rgb}{0.88,1,1}
\definecolor{LightGrey}{rgb}{0.88,0.88,0.88}
\newcommand{\demph}[1]{\textcolor{gray}{#1}}

\iccvfinalcopy 

\def\iccvPaperID{7200} 

\ificcvfinal\fi

\begin{document}

\title{MoMo: A shared encoder \textbf{Mo}del for text, image and multi-\textbf{Mo}dal representations}

\author{Rakesh Chada\\
Amazon Alexa AI\\
{\tt\small rakchada@amazon.com}
\and
Zhaoheng Zheng\footnote{Work completed during Zhaoheng’s internship at Amazon.}\\
University of Southern California\\
{\tt\small zhaoheng.zheng@usc.edu}
\and
Pradeep Natarajan\\
Amazon Alexa AI\\
{\tt\small natarap@amazon.com}
}
\maketitle
\ificcvfinal\thispagestyle{empty}\fi

\begin{abstract}
   We propose a self-supervised shared encoder model that achieves strong results on several visual, language and multimodal benchmarks while being data, memory and run-time efficient. We make three key contributions. First, in contrast to most existing works, we use a single transformer with all the encoder layers processing both the text and the image modalities. Second, we propose a stage-wise training strategy where the model is first trained on images, then jointly with unimodal text and image datasets and finally jointly with text and text-image datasets. Third, to preserve information across both the modalities, we propose a training pipeline that learns simultaneously from gradient updates of different modalities at each training update step. The results on downstream text-only, image-only and multimodal tasks show that our model is competitive with several strong models while using fewer parameters and lesser pre-training data. For example, MoMo performs competitively with FLAVA on multimodal (+3.1), image-only (+1.1) and text-only (-0.1) tasks despite having 2/5th the number of parameters and using 1/3rd the image-text training pairs. Finally, we ablate various design choices and further show that increasing model size produces significant performance gains indicating potential for substantial improvements with larger models using our approach.
\end{abstract}

\section{Introduction}

\label{sec:intro}

Self-supervised learning has proven to be a very effective training mechanism for learning representations that transfer well to various downstream tasks\cite{devlin2018bert,he2022masked,radford2021learning,baevski2022data2vec}. It has been shown to be successful across text, image, speech and video modalities. In a unimodal setting, inputs are fed into a single model, \eg a Transformer\cite{vaswani2017attention}, while features are learned through self supervision (such as masked input reconstruction) on large datasets \cite{devlin2018bert,liu2019roberta,bao2022beit,he2022masked}. In multimodal settings, specifically with vision and language, image and text features are typically processed by separate unimodal encoders, with a multimodal encoder on top, encoding the joint sequence to learn cross-modality interactions. Such multimodal models rely on either huge training corpora or models with an enormous number of parameters. For instance, CLIP \cite{radford2021learning} is trained on 200M image-text pairs, ALIGN \cite{jia2021scaling} and SimVLM \cite{wang2021simvlm} are trained on about 1B image-text pairs. Recently, FLAVA \cite{singh2022flava} has been proposed to reduce the reliance on large corpora while still serving a multitude of tasks effectively. However, FLAVA requires 75M text samples, 70M image-text pairs and an external image tokenizer trained with 250M image-text pairs \cite{ramesh2021zero}. Furthermore, for its model architecture, FLAVA leverages three different encoders for visual, linguistic and multimodal representations.

\begin{figure}
    \centering
    \includegraphics[width=\linewidth]{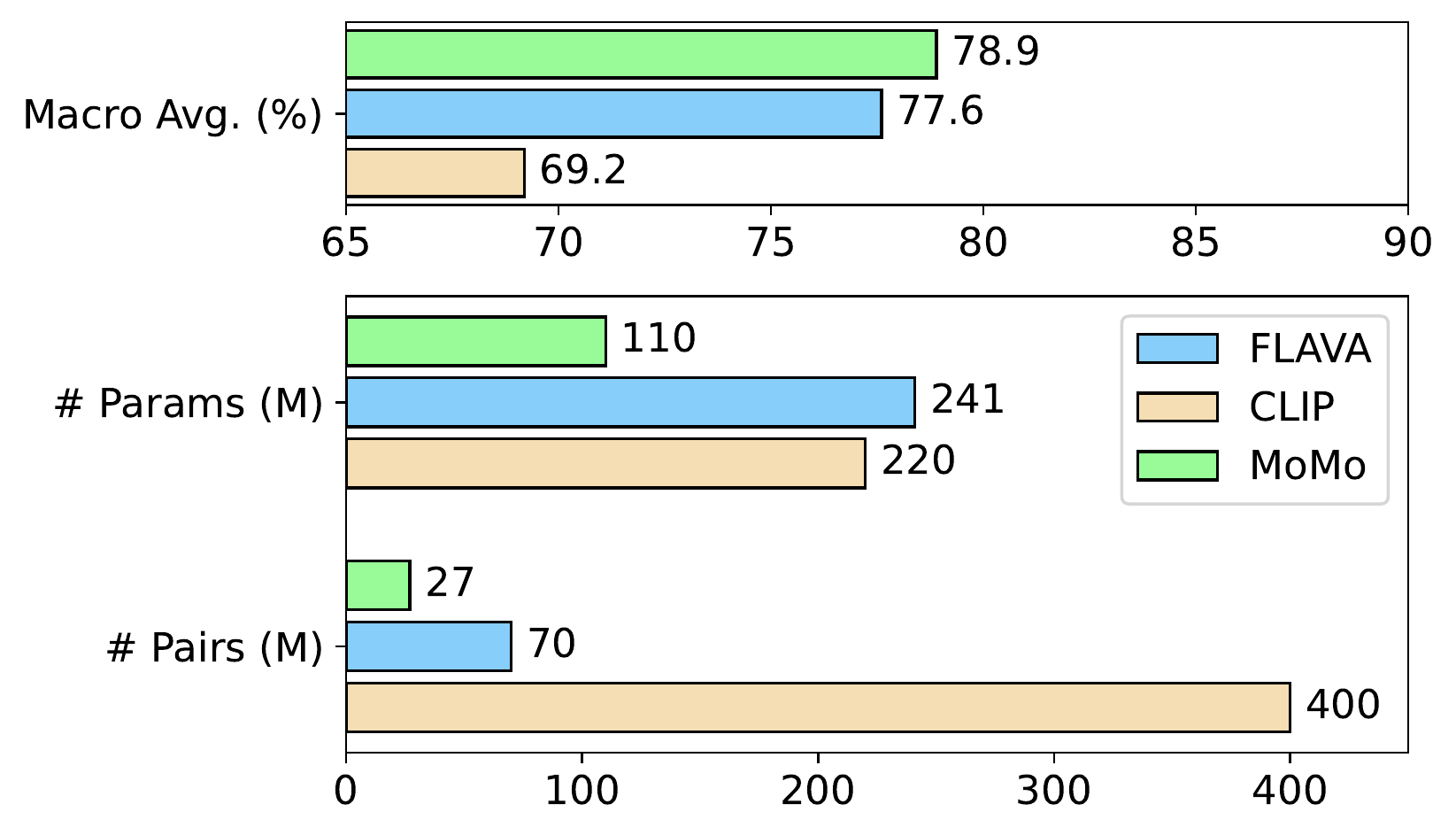}
    \caption{Comparison with FLAVA \cite{singh2022flava} and CLIP \cite{radford2021learning}: MoMo achieves the best macro average across vision, language and multimodal benchmarks, which is the mean of the average of all 3 task types, with $\approx$ 2/5th the parameters and 1/3rd the training image-caption pairs. }
    
    \label{fig:intro}
\end{figure}

\begin{figure*}[!h]
    \centering
    \includegraphics[width=\textwidth]{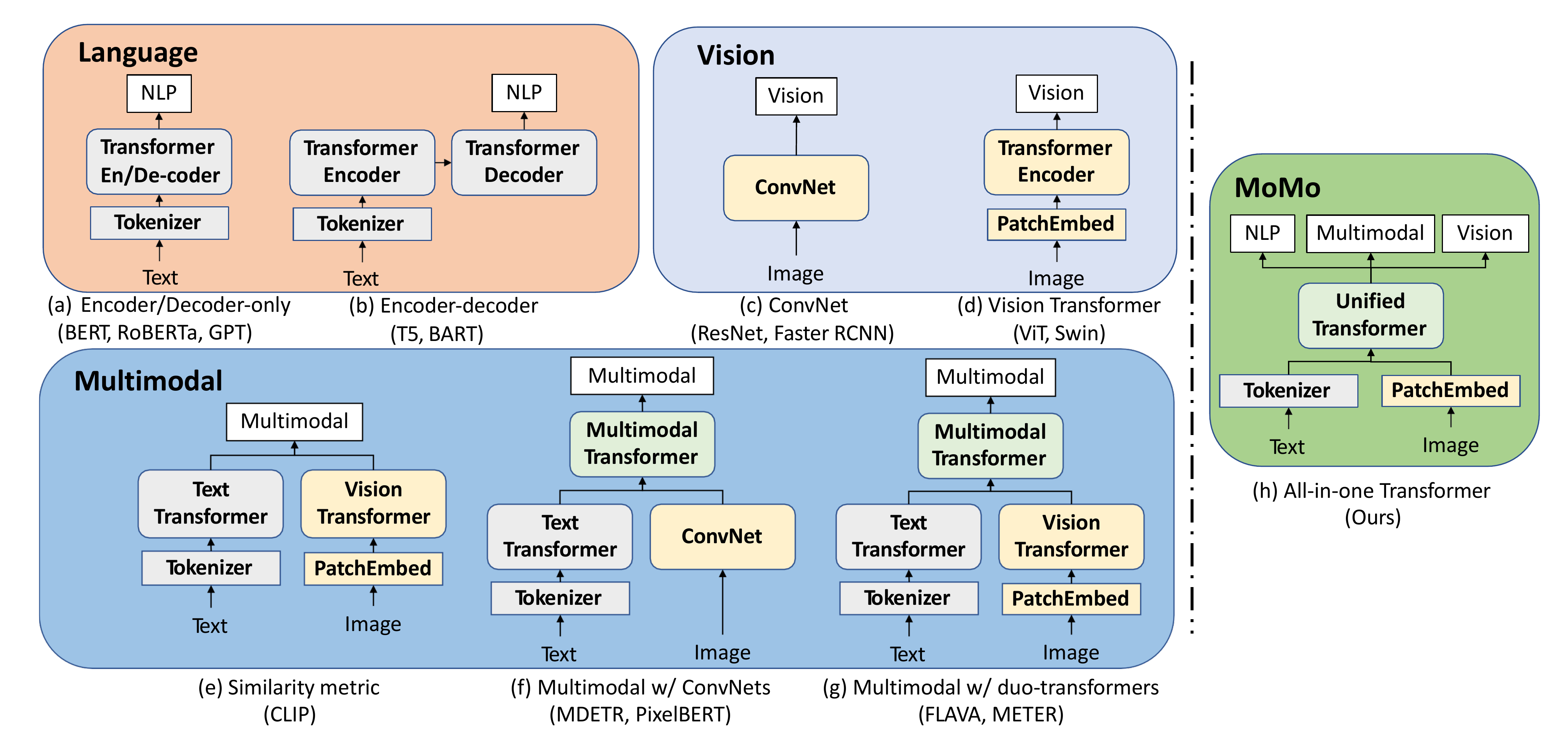}
    \caption{Illustrations of how models process input data across various modalities.  \textbf{Language tasks}: (a) Encoder-only architecture; (b) Encoder-decoder architecture. \textbf{Vision tasks}: (c) Convolutional Neural Networks; (d) Vision Transformers. \textbf{Multimodal tasks}: (e) Similarity-based metric; (f) Multimodal transformer with ConvNets; (g) Multimodal transformer with duo-transformers; (f) \textit{MoMo}: Multiple tasks across multiple modalities through an all-in-one unified transformer.}
    \label{fig:model-illustration}
\end{figure*}

In this work, we propose a simple unified model design where the same Transformer encoder processes both the unimodal vision/language inputs and the cross-modal vision+language inputs. Specifically, once the text tokens and the image patches are embedded into the latent space, they are processed through the same transformer encoder layers for all tasks. As shown in Fig.~\ref{fig:intro}, with about 1/3rd the dataset size and 2/5th the number of parameters, our model, MoMo outperforms FLAVA on macro average accuracy, which is computed over multiple vision, language and vision-language benchmarks. Due to the reduced model size and the shared parameters, our model is considerably faster during inference in comparison to other models (upto 35\% speedup for multimodal tasks such as Visual Question Answering). Furthermore, in addition to providing data efficiency and run-time efficiency, our model simplifies the production deployment pipelines by eliminating the need to maintain separate models for each of those modalities. This is enabled by its shared encoder structure, which handles all language, vision and multimodal tasks.

We also propose a new training strategy to address challenges in learning multimodal representations by the shared encoder. Since all the encoder parameters are shared across modalities, if the model is trained independently on datasets from different modalities, there tends to be information loss. To mitigate this, we design a multi-stage training pipeline where we jointly iterate through both the modalities during each training step (Section \ref{sec:pipeline}). We accumulate gradients from both the modalities and perform a single weight update. Our experiments show that such design leads to effective multimodal representation learning.

We use masked input prediction\cite{devlin2018bert,he2022masked,bao2022beit,xie2022simmim} objective for unimodal stages. For cross-modal learning, we mask both the modalities at higher fractions to enable rich cross-modal interactions. A key design feature of our model is that the encoder processes only the visible part of the inputs, regardless of their modality. This leads to a significantly improved training efficiency. 

The rest of the paper is organized as follows: in Section \ref{sec:related_work}, we present an overview of related work and compare MoMo to popular models. In Section \ref{sec:momo} we describe details training, model architecture, pre-training datasets and other implementation details. In Section \ref{sec:expts} we present experimental results and detailed ablation studies. Finally, in Section \ref{sec:conclusion} we present conclusions and future work.

\begin{table*}[t]
\centering
\footnotesize
\begin{tabular}{lccccccc}
\toprule
& \shortstack{Public \\Datasets?}& \shortstack{Types of \\inputs accepted} & \shortstack{External \\Image processing?} & \shortstack{\#Pairs \\ Image-Text}&\shortstack{\#Params \\ Text} & \shortstack{\#Params \\ Image} & \shortstack{\#Params \\ Multimodal} \\ 
\midrule
MoMo (ours) & \multirow{2}{*}{Yes} & \multirow{2}{*}{Text,Image,Multimodal} & \multirow{2}{*}{No} & \multirow{2}{*}{27M} & 110M & 86M & 110M \\
MoMo-Large (ours) & &  &  & & 335M & 303M & 335M \\
\midrule
FLAVA \cite{singh2022flava}& Yes & Text,Image,Multimodal & Image Tokenization (BEIT) & 70M & 110M & 86M & 241M\\
\midrule
CLIP \cite{radford2021learning} & No & Text,Image & No & 400M & 63M & 86M & -\\
\midrule
BEIT-3 \cite{wang2022image} & Yes & Text, Image,Multimodal & Image tokenization (BEIT) & 21M & 1.0B$^\dag$ & 1.0B$^\dag$ & 1.9B$^\dag$\\
\midrule
FLAMINGO \cite{alayrac2022flamingo} & No & Image,Multimodal & No & 2.1B & - & 1.4B$^\dag$ & 3.2B$^\dag$\\
\midrule
\multirow{2}{*}{LEMON \cite{hu2022scaling}} & \multirow{2}{*}{No} & \multirow{2}{*}{Multimodal} & Region features & \multirow{2}{*}{200M} & \multirow{2}{*}{-} & \multirow{2}{*}{-} & \multirow{2}{*}{110M}\\
& & & \& object detection features & & & &\\
\midrule
VLMo \cite{bao2021vlmo} & Yes & Text,Image,Multimodal & Image tokenization(BEIT) & 10M & 110M & 110M & 130M\\
\midrule
Coca \cite{yu2022coca} & No & Image, Multimodal & No & 4.8B & - & 86M & 383M\\
\midrule
UNiTER \cite{chen2020uniter} & Yes & Multimodal & Region features & 5M & - & - & 110M\\
\bottomrule
\end{tabular}

\caption{Comparison of various design choices across several strong multimodal models vs MoMo. BEIT tokenizer is an externally learnt VQ-VAE based model trained on 250 million image-text pairs. The number of image-text pairs is the size of pre-training data. Numbers of parameters are calculated out of the number of model parameters used for running inference on downstream tasks (retrieval, classification, VQA) listed in the paper. $\dag$: We report the size of the smallest model available in the original paper.}
\label{tab:datasets}
\end{table*}

\section{Related Work}
\label{sec:related_work}
\textbf{Language Transformers.} Transformers, introduced in Vaswani \etal \cite{vaswani2017attention}, have first proven to be successful for several Natural Language Processing (NLP) tasks. A variety of methods have been built on top of the transformer architecture with different objectives. Notably, BERT \cite{devlin2018bert}, with an encoder-only architecture in Fig.\ref{fig:model-illustration} (a), leverages the bidirectional attention mechanism and train the model with Masked Language Modeling (MLM) on large text corpus. RoBERTa \cite{liu2019roberta}, ELECTRA \cite{Clark2020ELECTRA:}, ALBERT \cite{lan2019albert}, and DeBERTa \cite{he2021deberta} introduced extensions to BERT for Language Modeling (LM). Other language transformer models \cite{brown2020language,lewis2020bart} adopt the encoder-decoder architecture with autoregressive objective and unidirectional attention.  

\textbf{Vision Transformers.} Prior to transformers, the prevailing paradigm for vision tasks is to process images through convolutional neural networks (ConvNets), as in Fig.~\ref{fig:model-illustration}(c). Inspired by BERT, Dosovitskiy \etal \cite{dosovitskiy2021an} treat an image as a sequence of image patches and propose to perform vision tasks through Vision Transformer (ViT), a BERT-like architecture. Numerous efforts have been made to improve the performance of ViT: DEiT \cite{touvron2021training} introduces distillation into the training phase; CaiT \cite{touvron2021going} explores deeper ViT structures, while VOLO\cite{yuan2022volo} designs a sliding window mechanism to aggregate local features; SwinTransformer\cite{liu2021swin} designs a hierarchical transformer structure with shifting windows. Furthermore, BEiT \cite{bao2022beit} tokenizes image patches through an external tokenizer and proposes to pre-train ViT with a BERT-style objective, named as Masked Image Modelling (MIM). He \etal \cite{he2022masked} shows that ViT better generalizes under pixel-level supervision with aggressive masking.

\textbf{Multimodal Transformers.} The success of transformer-based architectures in the field of NLP has sparked research efforts in multi-modality transformers, especially in the domain of Vision-Language Pre-training (VLP). Prior to ViT, models relied on external image processors to feed as inputs to transformer based models. \cite{jia2021scaling,tan2019lxmert,li2020does,chen2020uniter,su2019vl,li2020oscar,zhang2021vinvl,goenka2022fashionvlp} leverage external object detectors to generate input sequences of regional features. On the other hand, \cite{huang2020pixel,huang2021seeing,shen2021much,kamath2021mdetr,alayrac2022flamingo} convert feature grids from Convolutional Neural Networks (CNNs) into visual input sequences. Recent research \cite{radford2021learning,singh2022flava,yang2022vision,duan2022multi,dou2022empirical,kim2021vilt,akbari2021vatt} converts vision inputs into sequences through linear projection. Notably, CLIP \cite{radford2021learning} leverages contrastive loss to extract visual-linguistic embeddings from noisy web image-data; FLAVA \cite{singh2022flava} aims at unifying vision, language and multimodal tasks with one VLP model; SimVLM \cite{wang2021simvlm} encodes images through a CNN and feed inputs from both modalities into a encoder-decoder transformer. VATT \cite{akbari2021vatt} proposes a modality-agnostic transformer to align features from different modalities; VLMo \cite{bao2021vlmo} proposes to unify unimodal and multimodal tasks by a set of modality-specific experts but, unlike MoMo, it still requires particular modules for each modality.

 \begin{figure*}[t]
\centering
\includegraphics[width=0.95\linewidth]{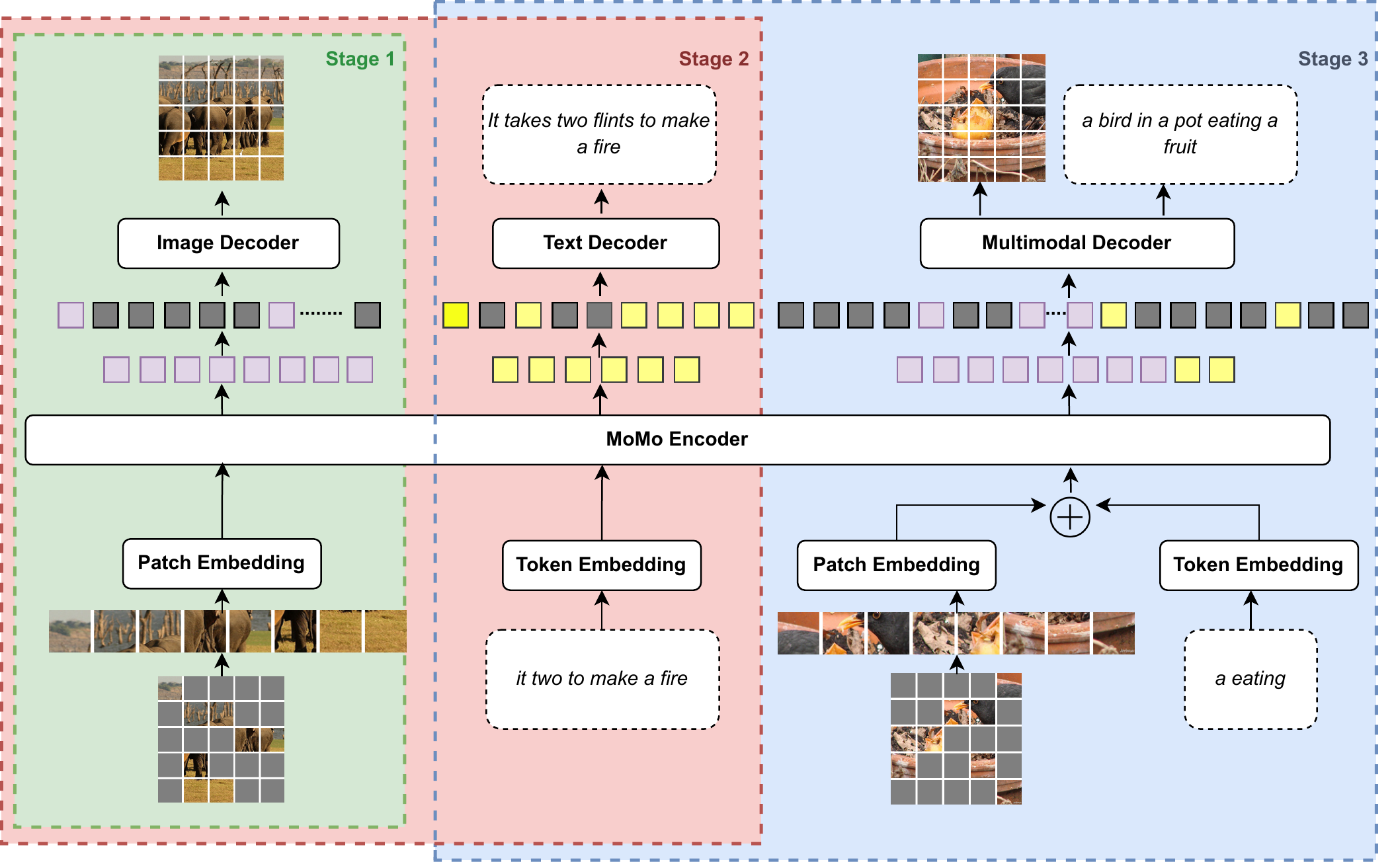}
\caption{The architecture and training stages of MoMo. The model is first trained with unimodal image datasets. Then simultaneously with unimodal image and unimodal text datasets. And finally, simultaneously, with unimodal text and multimodal image-text datasets. Each stage is executed sequentially and is initialized from the previous stage model's weights. For downstream task fine-tuning, the decoder is discarded and only the encoder weights are used.}
\label{fig:arch}
\vspace{-10pt}
\end{figure*}

The existing vision-language models can be broadly categorized into a single-stream or dual-stream architectures (Fig.\ref{fig:model-illustration}(e)(f)(g)). The dual-stream models have separate encoders for text and image (and typically also a separate multimodal encoder) while the single-stream models process them through a joint encoder. MoMo is a single-stream model. However, unlike most other single-stream models, it is optimized to do well across all of text-only, image-only and multimodal tasks (Table \ref{tab:datasets}). VLMO \cite{bao2021vlmo} deviates from these common architectural choices and uses modality-specific experts to better capture text, image and multi-modal features. Their work also studies a variation of this architecture that doesn't use these specific experts and thereby making the architecture closer to MoMo. However, this was only evaluated on multimodal and image-only tasks. MoMo studies training a shared encoder model to work well simultaneously on unimodal text and image tasks along with the multimodal tasks. Also, due to the shared encoder design, MoMo is computationally efficient during inference for multimodal tasks.

\section{The MoMo Framework}
\label{sec:momo}
Our goal is to build a compact and efficient model that is capable of handling both unimodal and multimodal tasks. We show that this is possible through the training pipeline design described in Section ~\ref{sec:pipeline} and modeling architecture choices in Section ~\ref{sec:model}.

\subsection{Training Objectives}
The loss functions used during different stages of training MoMo are described below:

\textbf{Masked Image Modeling (MIM).} Following \cite{he2022masked}, we adopt an encoder-decoder architecture during training and ask the model to reconstruct patches that are randomly masked out, at the pixel level. During training, we remove masked patches at the encoding stage and restructure the input sequence with \textbf{\texttt{[MASK]}} tokens before the reconstruction. As shown in Eqn.~\ref{eqn:mim}, where $x$ and $\hat x$ are original and reconstructed pixels, We use Mean Square Error (MSE) to measure the loss between the reconstructed image and the original image.
\begin{equation}
    \mathcal{L}_{MIM} = ||x - \hat{x}||^2_2
    \label{eqn:mim}
\end{equation}
\textbf{Masked Language Modeling (MLM).} Similar to \cite{devlin2018bert,liu2019roberta}, we adopt MLM when training MoMo on text datasets. We randomly mask out a portion of the tokens of the input sequence and ask the model to reconstruct the missing tokens. We utilize the cross-entropy loss, elaborated in Eqn.~\ref{eqn:mlm} to measure the disparity between predictions and targets. The difference between our approach and \cite{devlin2018bert,liu2019roberta} is that we follow the encoder-decoder structure in \cite{he2022masked}, where masked tokens are removed for the encoder and are reconstructed through a separate decoder. This modification accelerates the training process as the length of the input sequence of the encoder is much shorter.
\begin{equation}
    \mathcal{L}_{MLM} = \sum_{x \in X_{M}} -\log P(\hat x = x)
    \label{eqn:mlm}
\end{equation}
\textbf{Cross-modal masking Loss (CMM). } To enhance interactions between the modalities (Fig.~\ref{fig:momo-vis}), we use a cross-modal masking loss on the concatenated image-text sequence. Specifically, both the image tokens ($M_I$) and text tokens ($M_T$) are masked at 75\% and the model is asked to reconstruct the masked tokens using visible tokens from both the modalities. 
\begin{equation}
    \mathcal{L}_{CMM} = \sum_{x \in X_{M_I}} \mathcal{L}_{MIM} + \sum_{x \in X_{M_T}} \mathcal{L}_{MLM}
    \label{eqn:cmm}
\end{equation}
\textbf{Global Contrastive Loss. } To align image-text representations, we also include an image-text contrastive loss \cite{radford2021learning, singh2022flava}, shown in Eqn.~\ref{eqn:gc}, along with the masked token reconstruction loss during stage 3. Similar to FLAVA, we use a global contrastive loss where the image-text pairs from all GPU nodes are accumulated before the loss calculation.
\begin{equation}
\mathcal{L}_{GC} = \sum_{i} -\log \frac{\exp(s(\mathbf{f}^{(i)}_{img}, \mathbf{f}^{(i)}_{text}) / \tau)}{\sum_{j} \exp(s(\mathbf{f}^{(i)}_{img}, \mathbf{f}^{(j)}_{text}) / \tau)}
\label{eqn:gc}
\end{equation}
\textbf{Image-Text Matching Loss. } To further enhance cross-modality interactions, we also use Image-text matching loss following previous works \cite{singh2022flava} . Specifically, we collect positive and negative image-text pairs in each batch guided by similarity scores from the contrastive learning \cite{li2021align}. Then, we apply a classifier on top of the \textbf{\texttt{[MASK]}} token to predict if the image-text pair match.

\subsection{Multi-Stage Training}
\label{sec:pipeline}
A key component of our training pipeline is learning simultaneously from multiple modalities. 

To effectively train MoMo with multiple modalities, we develop a three- stage training pipeline as illustrated in Figure \ref{fig:arch}:

\textbf{Stage 1}: The model is trained on the unimodal image data with MIM loss with a masking ratio of 75\%.

\textbf{Stage 2}: After stage 1, the model is further trained simultaneously on unimodal image data and text data. For image, masked patch reconstruction loss with 75\% masking is used and for text, masked token reconstruction loss with 15\% masking is used.

\textbf{Stage 3}: The model is initialized with the weights from stage 2 and is trained simultaneously on unimodal text data and multimodal image-text data. For unimodal text data, masked token reconstruction loss with 15\% masking is used. For multimodal image-text data, a combination of cross-modal masking, global contrastive loss and image-text matching losses are used.

\subsection{Model Architecture}
\label{sec:model}
As illustrated in Figure \ref{fig:arch}, our model is a transformer-based structure containing an encoder and a shallow decoder. We follow the design of ViT for both encoder and decoder. We use separate decoders for each stage as we found this to be effective while jointly training on image and text modalities. We discuss the effect of separate decoders in Sec.\ref{sec:ablate}. During downstream fine-tuning, all decoders are discarded and only the encoder weights are updated. Our model, MoMo, consists of 12 encoder layers with a hidden dimension of 768 and 8 decoder layers with a hidden dimension of 512. The token embedding weights and the language modeling output head weights are tied. Following MAE~\cite{he2022masked}, the encoder processes only the visible part of the inputs for all stages.

\subsection{Datasets}
\label{sec:dataset}
We use ImageNet-1K \cite{deng2009imagenet} for Stage 1 training and  Wikipedia \& Books Corpus for Stage 2 training. For Stage 3, we use a subset of the Public Multimodal Dataset (PMD) from FLAVA \cite{singh2022flava}. This subset consists of COCO \cite{lin2014microsoft}, SBU Captions \cite{ordonez2011im2text}, Conceptual Captions \cite{sharma2018conceptual,changpinyo2021conceptual}, Visual Genome \cite{krishna2017visual} and RedCaps \cite{desai2021redcaps} datasets. We call this dataset as PMD-Subset. The text samples in PMD-Subset have an average length of $\approx$12 tokens. Statistically, we have 6.6M documents for unimodal language tranining, 1M images for unimodal vision training and 26.6M image-text pairs for multimodal training. For fair comparison, we follow the data filtering process from \cite{chen2020uniter} to exclude validation and test images that appear in downstream tasks.



\begin{table*}[t]
\setcounter{magicrownumbers}{0}
\footnotesize
\centering
\setlength{\tabcolsep}{5pt}
\begin{tabular}{ccl|c@{\ \ }c@{\ }c|c@{\ \ }c@{\ \ }ccccc|c}
\toprule
& \multicolumn{1}{@{}c@{}}{public}& & \multicolumn{3}{c|}{Multimodal Tasks} & \multicolumn{7}{c|}{Language Tasks} & ImageNet \\
& \multicolumn{1}{@{}c@{}}{data}& & VQAv2 & COCO & Flickr30K & SST-2 & RTE & MRPC & QQP & MNLI & QNLI & STS-B & linear eval \\
\midrule
\demph{\texttt{\rownumber}} & \demph{\checkmark}& \demph{BERT$_\text{base}$} \cite{devlin2018bert} & \demph{--} & \demph{--} & \demph{--} &  \demph{92.5} & \demph{62.5} & \demph{81.9/87.6} & \demph{90.6/87.4} & \demph{84.4} & \demph{91.0} & \demph{88.1}  & \demph{--} \\
\midrule
\texttt{\rownumber}  & \xmark & CLIP-ViT-B/16 \cite{radford2021learning} & 55.3 & 55.2 & \underline{81.6} & 88.2 & 55.2 & 74.9/65.0 & 76.8/53.9 & 33.5 & 50.5 & 16.0 & 80.2 \\
\texttt{\rownumber} & \xmark&SimVLM$_\text{base}$ \cite{wang2021simvlm} & \underline{77.9} & -- & --  & 90.9 & \underline{63.9} & 75.2/84.4 & \underline{90.4/87.2} & \underline{83.4} & \underline{88.6} & -- & \underline{80.6} \\
\midrule
\texttt{\rownumber} & \checkmark&VisualBERT \cite{li2020does} & 70.8 & -- & -- & 89.4 & 56.6 & 71.9/82.1 & 89.4/86.0 & \bf  81.6 & 87.0 & 81.8 & -- \\
\texttt{\rownumber} & \checkmark&UNITER$_\text{base}$ \cite{chen2020uniter} & 72.7 & -- &-- & 89.7 & 55.6 & 69.3/80.3 & 89.2/85.7 & 80.9 & 86.0 & 75.3 & -- \\
\texttt{\rownumber} & \checkmark&VL-BERT$_\text{base}$ \cite{su2019vl} & 71.2 & -- & --~  & 89.8 & 55.7 & 70.6/81.8 & 89.0/85.4 & 81.2 & 86.3 &  82.9 & -- \\
\texttt{\rownumber} & \checkmark&ViLBERT \cite{lu2019vilbert} & 70.6 & -- & --  & 90.4 & 53.7 & 69.0/79.4 & 88.6/85.0 & 79.9 & 83.8  & 77.9 & -- \\
\texttt{\rownumber} & \checkmark&LXMERT \cite{tan2019lxmert} & 72.4 & -- & --~  & 90.2 & 57.2 & 69.7/80.4 & 75.3/75.3 & 80.4 & 84.2 & 75.3 & -- \\
\texttt{\rownumber} & \checkmark&UniT \cite{hu2021unit} & 67.0 & --  & --~  & 89.3 & -- & -- & 90.6/ ~~--~~ & 81.5 & \bf  88.0 & -- & -- \\
\texttt{\rownumber} & \checkmark&CLIP-ViT-B/16 (PMD) \cite{singh2022flava} & 59.8 & 50.6 & 72.5 & 83.5 & 53.1 & 63.5/68.7 & 75.4/43.0 & 32.9 & 49.5 & 13.7 & 73.0 \\
\texttt{\rownumber} & \checkmark&FLAVA \cite{singh2022flava} & \bf 72.8 & 56.3 & \textbf{79.1}  & \underline{\bf 90.9} & 57.8 & \underline{\bf 81.4}/86.9 & \underline{\bf 90.4/87.2} &  80.3 & 87.3 & 85.7 & 75.5 \\
\rowcolor{LightGrey}
\texttt{\rownumber} & \checkmark&MoMo (ours) & 71.2 & \underline{\textbf{64.0}} & 78.2 &  89.6 &   \textbf{59.6} & 81.3/\underline{\textbf{87.0}} & 90.2/86.6 &  78.1 & 86.0 & \underline{\bf  87.0} & \bf  75.7 \\
\midrule
\texttt{\rownumber} & \checkmark&MoMo-Large (ours) & 75.0 & 72.0 & 84.1 &  90.4 &   63.9 & 82.8/88.0 & 90.7/87.7 & 80.6 & 88.0 & 88.0/87.8 & 79.1 \\
\bottomrule
\end{tabular}
\caption{Comparing MoMo to previous models on multimodal, language, and image tasks. We report results on dev sets of the GLUE benchmark \cite{wang-etal-2018-glue}. We report accuracy/F1 for MRPC and QQP; the Pearson/Spearman correlation for STS-B; Averaged recall at Top-1/5 for zero-shot retrieval on COCO and Flickr30K; test-dev VQA score for VQAv2 and accuracy for all the other tasks. Results of BERT and other methods are taken from \cite{singh2022flava}. Note that SimVLM, CLIP and FLAVA are pretrained on much larger datasets than MoMo (1.8B, 400M \& 75M vs 27M pairs).
\textbf{Bold} signifies the best result on public data while \underline{underlined} indicates the overall best result.}
\vspace{-10pt}
\label{tab:sota_comp}
\end{table*}

\subsection{Simultaneous learning}
\label{sec:cmga}
Apart from the three-stage pipeline, another key ingredient of MoMo is training simultaneously on multiple modalities. Specifically, during training, we simultaneously iterate through the image and text datasets and learn from both the modalities during each training step. For gradient update, we propose to accumulate the gradients for both the modalities and perform a single update per training step. We refer to it as Cross-Modality Gradient Accumulation (CMGA).

\subsection{Implementation details}
For all stages, We use 64 V100 GPUs to train the model. During all stages, we append a [CLS] token at the beginning of the sequence. This is used for computation during contrastive loss and classification tasks.

\textbf{Stage 1}: We follow settings from MAE \cite{he2022masked}. Specifically, the model is trained for 1600 epochs on ImageNet dataset with a base learning rate (LR) of 1.5e-4, batch size of 64, linear warmup of 40 epochs followed by a cosine decay. We use the Random resizing and cropping as the only augmentation. For our default model, we re-use weights from the publicly available MAE model. 

\textbf{Stage 2}: The dataloader during this stage iterates simultaneously through unimodal text and image datasets. We repeat the imagenet data five times during a single epoch so the image data loader roughly matches the text data loader's size. The model is trained for 100 epochs. The learning rate is set to 5e-5 with a linear warmup for 10 epochs followed by cosine decay until the end of training. The per-device batch size is set to 64.

\textbf{Stage 3}: The model is trained for 100 epochs without any data repetition at each epoch. Since the text and multimodal datasets are iterated through simultaneously, an "epoch" here covers roughly 1/4th of the multimodal dataset (the same size as the unimodal text dataset). For image augmentation, we use random resize/cropping and RandAugment \cite{NEURIPS2020_d85b63ef}. The learning rate is set to 1e-4 with a linear warmup for 10 epochs followed by cosine decay. The per-device batch size is 64.\\

\begin{table}[t]
\centering
\footnotesize
\resizebox{\linewidth}{!}{%
\begin{tabular}{l@{\ }c|cc|c}
\toprule
&  & \multicolumn{1}{c}{MoMo} & \multicolumn{1}{c}{$\mathrm{FLAVA}$} & \multicolumn{1}{c}{CLIP} \\
\midrule 
Datasets & Eval. & PMD-Sub & PMD & 400M\cite{radford2021learning} \\
\midrule 
NLP Active Param. & & 110M & 110M & 63M\\
\midrule
MNLI \cite{williams2018broad} & FT & 78.1 & \underline{\textbf{80.3}} & 33.5 \\
MRPC \cite{dolan2005automatically} & FT & \underline{\textbf{84.2}} & 84.2 & 69.9 \\
QQP \cite{qqp} & FT & 88.4 & \underline{\textbf{88.7}} & 65.3 \\
SST-2 \cite{socher2013recursive} & FT & 89.6 & \underline{\textbf{90.9}} & 88.2 \\
QNLI \cite{rajpurkar2016squad} & FT & 86.0 & \underline{\textbf{87.3}} & 50.5 \\
RTE \cite{dagan2005pascal,bentivogli2009fifth} & FT & \underline{\textbf{59.6}} & 57.8 & 55.2 \\
STS-B \cite{agirre2007proceedings} & FT & \underline{\textbf{87.0}} & 85.7 & 16.0 \\
\midrule 
\textbf{NLP Avg.} &  & 81.8 & \underline{\bf 82.1} & 54.1 \\
\midrule 
Vision Active Param. & & 86M & 86M & 86M\\
\midrule 
ImageNet \cite{deng2009imagenet} & LE & \textbf{75.7} & 75.5 & \underline{80.2} \\
Food101 \cite{bossard2014food} & LE & \textbf{90.7} & 88.5 & \underline{91.6} \\
CIFAR10 \cite{krizhevsky2009learning} & LE & \underline{\textbf{95.0}} & 92.8 & \underline{94.9} \\
CIFAR100 \cite{krizhevsky2009learning} & LE & \textbf{80.5} & 77.7 & \underline{81.1} \\
Cars \cite{krause20133d} & LE & \textbf{71.6} & 70.9 & \underline{86.0} \\
Aircraft \cite{maji2013fine} & LE & \textbf{51.2} & 47.3 & \underline{51.4} \\
DTD \cite{cimpoi2014describing} & LE & \underline{\textbf{79.7}} & 77.3 & 78.5 \\
Pets \cite{parkhi2012cats} & LE & \underline{\textbf{91.7}} & 84.8 & 91.7 \\
Caltech101 \cite{fei2004learning} & LE & {92.2} & \underline{\textbf{95.7}} & 95.5 \\
Flowers102 \cite{nilsback2008automated} & LE & \underline{\textbf{97.8}} & 96.4 & 97.1 \\
MNIST \cite{lecun-mnisthandwrittendigit-2010} & LE & 97.0 & 98.4 & \underline{99.0} \\
STL10 \cite{coates2011analysis} & LE & 98.2 & \textbf{98.9} & \underline{99.1} \\
EuroSAT \cite{helber2019eurosat} & LE & {95.8} & \underline{\textbf{97.3}} & 95.4 \\
GTSRB \cite{stallkamp2011german} & LE & \textbf{80.7} & 79.5 & \underline{88.6} \\
SST \cite{radford2021learning} & LE & 56.7 & \bf 57.1 & \underline{74.7} \\
\midrule 
\textbf{Vision Avg.} &  & \bf 83.6 & 82.5 & \underline{86.7} \\
\midrule 
Multimodal Active Param. & & 110M & 241M & 220M\\
\midrule
VQAv2 \cite{balanced_vqa_v2} & FT & 71.2 & \underline{\textbf{72.5}} & 54.8 \\
Flickr30K \cite{plummer2015flickr30k} TR@1 & ZS & \textbf{74.2} & 67.7 & \underline{82.2} \\
Flickr30K \cite{plummer2015flickr30k} TR@5 & ZS & 93.7 & \textbf{93.2} & \underline{96.6} \\
Flickr30K \cite{plummer2015flickr30k} IR@1 & ZS & 60.9 & \underline{\textbf{65.2}} & 62.1 \\
Flickr30K \cite{plummer2015flickr30k} IR@5 & ZS & 84.8 & \underline{\textbf{89.4}} & 85.7 \\
COCO \cite{lin2014microsoft} TR@1 & ZS & \underline{\textbf{59.3}} & 42.1 & 52.5 \\
COCO \cite{lin2014microsoft} TR@5 & ZS & \underline{\textbf{84.1}} & 70.4 & 76.7 \\
COCO \cite{lin2014microsoft} IR@1 & ZS & \underline{\textbf{42.7}} & 38.4 & 33.1 \\
COCO \cite{lin2014microsoft} IR@5 & ZS & \underline{\textbf{70.6}} & 67.5 & 58.4 \\
\midrule
\textbf{Multimodal Avg.} &  & \underline{\textbf{71.3}} & 68.2 & 66.9 \\
\bottomrule

\end{tabular}
}

\caption{Performance comparison between MoMo, FLAVA and CLIP. The numbers for MRPC and QQP are the averages of accuracy and F1 for. For STS-B, it's Matthews correlation. The numbers for CoCo and Flickr30K are from top-1/5 zero-shot text and image retrieval. For other tasks, we report accuracy. \textbf{Bold} signifies the best result on public data while \underline{underlined} indicates the overall best result. Active parameters are the number of model parameters used during the forward pass for a task. FT, LE, and ZS stands for Fine-Tuning, Linear Eval and Zero-Shot, respectively.}
\vspace{-10pt}
\label{tab:flavaclip}
\end{table}

\vspace{-15pt}
\section{Experiments}
\label{sec:expts}
In this section, we perform quantitative evaluations over a span of downstream tasks along with ablation study.

\subsection{Quantitative Evaluation}

\textbf{Comparison with state-of-the-art VLP models.}
We compare the full MoMo model with several VLP models on multimodal tasks, language tasks and ImageNet linear probing. Results are reported in Tab.~\ref{tab:sota_comp}. MoMo either matches or outperforms most of the multi-modal models on majority of the tasks. Notably, MoMo surpasses other baselines, including CLIP, which is trained with 20X data than MoMo.

\textbf{In-depth comparison with FLAVA and CLIP.} We further compare our model in-depth with FLAVA and CLIP over a broader spectrum of tasks. Results are reported in Tab.~\ref{tab:flavaclip}. For evaluation protocol, we follow the settings in \cite{singh2022flava}, including Fine-Tuning (FT), Linear Evaluation (LE) and Zero-Shot (ZS). For zero-shot, models are evaluated on tasks without training. For linear evaluation, we freeze model weights and train an extra linear layer on top of it. For text and image retrieval tasks, MoMo's encoder is used as both the text and image encoder.

On unimodal NLP tasks, MoMo matches FLAVA performance despite using a shared encoder for both modalities. MoMo remains competitive in each of the individual benchmarks. On unimodal vision tasks, MoMo surpasses FLAVA on more than half of the selected benchmarks and remains comparable on others, achieving an average accuracy of 83.6\%, 1.1\% higher than FLAVA. 

On multimodal tasks, MoMo performs better on average (+3.1\%) than the state-of-the-art multimodal model FLAVA. Performance gains are especially prominent on COCO retrieval as MoMo improves over FLAVA by 17.2\% and 4.3\% for TR@1 and IR@1 respectively. However, the 110M parameter MoMo lags behind on VQA and Flickr30K Image retrieval. We hypothesize this gap could be further reduced by additionally pre-training MoMo on datasets of same scale as FLAVA. Furthermore, the slightly lower VQA score seems reasonable since MoMo uses 2/5th the number of parameters as FLAVA for this task.

\textbf{Increasing Model Size.} We conduct a larger-scale experiment where we replace VIT-B backbone with VIT-L and train on same data. The resulting model, MoMo-Large, contains 335M parameters. As seen in Tab. ~\ref{tab:sota_comp}, scaling up the model leads to remarkable performance gains across all tasks. On VQAv2, MoMo-Large improves our base model by 3.8\%, while improvements on image-text retrieval are 8.0\% and 5.9\% for COCO and Flickr30K, respectively. We also observe broad improvements over language tasks, while the performance on ImageNet is improved by 3.4\%.

\begin{table}[t]
\centering
\footnotesize
\begin{tabular}{rl@{\ }cccc}
\toprule
Modality & Datasets & Eval. & \multicolumn{1}{c}{Stage 1} & \multicolumn{1}{c}{Stage 2} & \multicolumn{1}{c}{Stage 3} \\
\midrule 
\multirow{7}{*}{Language} & MNLI \cite{williams2018broad} & FT & 59.0 & \textbf{78.5} & 78.1 \\
 & MRPC \cite{dolan2005automatically} & FT & 74.7 & \textbf{84.7} & 84.2 \\
 & QQP \cite{qqp} & FT & 67.4 & \bf 88.4 & \textbf{88.4} \\
& SST-2 \cite{socher2013recursive} & FT & 80.1 & 89.4 & \bf 89.6 \\
& QNLI \cite{rajpurkar2016squad} & FT & 61.5 & \textbf{86.7} & 86.0 \\
 & RTE \cite{dagan2005pascal,bentivogli2009fifth} & FT & 52.7 & 59.2 & \textbf{59.6 }\\
 & STS-B \cite{agirre2007proceedings} & FT & 7.4 & 86.2 & \textbf{87.0} \\
\midrule 
Vision & ImageNet \cite{deng2009imagenet} & LE & 66.9 & 52.9 & \bf 75.7 \\
\midrule 
\multirow{3}{*}{Multimodal} & VQAv2 \cite{balanced_vqa_v2} & FT & 56.7 & 57.1 & \bf 71.2 \\
 & Flickr30k \cite{balanced_vqa_v2} & ZS & - & - & \bf 78.2 \\
 & COCO \cite{balanced_vqa_v2} & ZS & - & - & \bf 64.0 \\
\bottomrule
\end{tabular}
\caption{Performance after different training stages in MoMo. Stages 2 and 3 bring in considerable performance gains for vision and multimodal tasks respectively.}
\label{tab:ablations}
\end{table}

\subsection{Ablation Study}
\label{sec:ablate}

In this section, we conduct ablation study over different model components to better understand MoMo.

\begin{table}[t]
\centering
\footnotesize
\begin{tabular}{rl@{\ }cccr}
\toprule
& & & \multicolumn{3}{c}{Last Two Training Stages} \\
\midrule
Modality & Dataset & Eval. & \multicolumn{1}{c}{Combined} & \multicolumn{1}{c}{Separate (\textit{$\Delta$})} \\
\midrule 
\multirow{7}{*}{Language} & MNLI \cite{williams2018broad} & FT & 74.0 & \textbf{76.5} (\textit{+2.5})\\
 & MRPC \cite{dolan2005automatically} & FT & 76.6 & \textbf{79.5} (\textit{+2.9}) \\ 
 & QQP \cite{qqp} & FT & 86.5 & \textbf{87.8}  (\textit{+1.3})\\ 
 & SST-2 \cite{socher2013recursive} & FT & 87.6 & \textbf{89.4} (\textit{+1.8})\\
 & QNLI \cite{rajpurkar2016squad} & FT & 82.4 & \textbf{85.0} (\textit{+2.6}) \\
 & RTE \cite{dagan2005pascal,bentivogli2009fifth} & FT & 53.4 & \textbf{57.4} (\textit{+4.0})\\
 & STS-B \cite{agirre2007proceedings} & FT & 71.2 & \textbf{83.0} (\textit{+11.8})\\
\midrule 
Vision & ImageNet \cite{deng2009imagenet} & LE & \bf 70.9 & 69.4 (\textit{-1.5})\\
\midrule 
\multirow{3}{*}{Multimodal}  & VQAv2 \cite{balanced_vqa_v2} & FT & \textbf{69.8} & 68.4 (\textit{-1.4})\\
 & Flickr30k \cite{balanced_vqa_v2} & ZS & \textbf{68.9} & 66.9 (\textit{-2.0}) \\
& COCO \cite{balanced_vqa_v2} & ZS & \textbf{54.5} & 53.1 (\textit{-1.4}) \\
\bottomrule
\end{tabular}
\caption{Performance comparison between combined and separate stages 2 and 3. Both the model variations are trained for 50k steps in this experiment. We observe that separating stage 2 and 3 results in better overall performance.}
\vspace{-10pt}
\label{tab:stagecombineablation}
\end{table}

\begin{table*}[t]
\centering
\footnotesize
\resizebox{\linewidth}{!}{
\begin{tabular}{rl@{\ }cccccccccc}
\toprule
&  & & \multicolumn{4}{c}{Stage 2} & \multicolumn{2}{c}{Stage 3} & \multicolumn{2}{c}{CMGA$^*$} \\
\midrule
 \multirow{2}{*}{Modality}& \multirow{2}{*}{Dataset} & \multirow{2}{*}{Eval.} & \multicolumn{2}{c}{Has simultaneous training?} & \multicolumn{2}{c}{Shared decoder?} & \multicolumn{2}{c}{Has simultaneous training?}& \multirow{2}{*}{\xmark} & \multirow{2}{*}{\cmark  (\textit{$\Delta$})} \\
& & & Yes & No (\textit{$\Delta$}) & Yes & No (\textit{$\Delta$}) & Yes & No (\textit{$\Delta$}) \\
\midrule
\multirow{7}{*}{Language} & MNLI \cite{williams2018broad} & FT & 78.5 & \textbf{78.7} \textit{(+0.2)} & 77.5 & \textbf{78.5 }(+1.0)  & 78.1 & \textbf{78.7} (+0.6) & 75.4 & \textbf{75.6} (\textit{+0.2}) \\
 & MRPC \cite{dolan2005automatically} & FT & \bf 84.7 & 83.3 (\textit{-1.4}) & \bf 86.4 & 84.7 (\textit{-1.7}) & \bf 84.2 & 78.3 (\textit{-5.9}) & 77.6 & \textbf{78.6} (\textit{+1.0}) \\
 & QQP \cite{qqp} & FT & \bf 88.3 & 88.1 \textit{(-0.2)} & 87.8 & \textbf{88.4} (\textit{+0.6}) & \bf 88.4 & 88.0 (\textit{-0.4}) &  87.2 & \textbf{87.7} (\textit{+0.5}) \\
 & SST-2 \cite{socher2013recursive} & FT & 89.4 &\textbf{90.1} (\textit{+0.7}) & \bf 89.4 & \textbf{89.4} (+\textit{0.0}) & \bf 89.6 & 88.5 (\textit{-1.1}) & 87.6 & \textbf{89.2} (\textit{+1.6}) \\
 & QNLI \cite{rajpurkar2016squad} & FT & 86.7 & \textbf{88.0} (\textit{+1.3}) & 86.0 & \textbf{86.7} (\textit{+0.7}) & \bf 86.0 & 84.2 (\textit{-1.8}) & 84.1 & \textbf{85.9} (\textit{+1.8}) \\
 & RTE \cite{dagan2005pascal,bentivogli2009fifth} & FT & 59.2 & \textbf{61.0} (\textit{+1.8}) & 58.4 & \textbf{59.2} (\textit{+0.8}) & \bf 59.6 & 54.1 (\textit{-5.5}) & 55.9 & \textbf{56.6} (\textit{+0.7})\\
 & STS-B \cite{agirre2007proceedings} & FT &  \bf 86.2 & 84.5 (\textit{-1.7}) &  84.7 & \textbf{86.2} (\textit{+1.5}) & \bf 87.0 & 82.5 (\textit{-4.5}) & 84.3 & \textbf{85.0} (\textit{+0.7}) \\
\midrule
Vision & ImageNet \cite{deng2009imagenet} & LE & \bf 52.9 & 9.5 (\textit{-43.4}) & 51.8 & \textbf{52.9} (\textit{+1.1}) & \bf 75.7 & 73.6 (\textit{-2.1}) & 71.9 & \textbf{72.2} (\textit{+0.3}) \\
\midrule
\multirow{3}{*}{Multimodal} & COCO \cite{deng2009imagenet} & ZS & - & - & - & - & 62.0 & \textbf{64.6} (\textit{+2.6}) & 54.1 & \textbf{55.4} (\textit{+1.3}) \\
 & Flickr30K \cite{deng2009imagenet} & ZS & - & - & - & - & \bf 78.7 & 78.1 (\textit{-0.6}) & 67.4 & \textbf{70.5} (\textit{+3.1}) \\
 & VQAv2 \cite{deng2009imagenet} & FT & - & - & - & - & 71.2 & \textbf{71.5} (\textit{+0.3}) & 69.3 & \textbf{70.2} (\textit{+0.9}) \\
\bottomrule
\end{tabular}
}
\caption{Ablation study of multiple modules inside MoMo. *Models under CMGA are trained for 50K steps.}
\vspace{-10pt}
\label{tab:main-ablate}
\end{table*}

\textbf{Multi-Stage Training.} Earlier work \cite{kim2021vilt} \cite{tan-bansal-2019-lxmert} reveal that initializing multimodal model with vision weights is more effective than initializing it with pre-trained text weights. On similar lines, we first train MoMo with unimodal image data before learning unimodal text and multimodal image-text representations. The performance of MoMo after each stage on various tasks is reported in Table \ref{tab:ablations}. After stage 1, the model achieves feasible image classification performance, while results on other tasks remain suboptimal. Stage 2 helps the model to learn language-related information and thus improves the performance on language tasks. Stage 3 further enhances the model capacity on multimodal tasks. We note that the performance on image classification degrades after Stage 2. We attribute this drop to the shared encoder design on learning different modalities. However, we observe that this drop is recovered and further improved during Stage 3, while language performance is retained. During stage 3, we also evaluated a variant that includes image-only training (along with text-only and multimodal). However, this didn't show significant differences on downstream tasks and hence we decided to save compute and skip this additional objective.

\textbf{Combining Stage 2 and 3.} We conduct experiments to investigate whether stage 2 and 3 can be combined. In the combined setting, the model is trained with the unimodal image, unimodal text and multimodal image-text losses at each training update step. We report the results in  Tab.~\ref{tab:stagecombineablation}. We observe that when stages 2 and 3 are combined, the performance of language tasks drops considerably (-3.8 points on average). The performance on image and multimodal tasks shows some improvement (+1.4 on average).

\begin{figure}[!t]
    \centering
    \footnotesize
    \def\lw{0.24}
    \def\llw{0.48}
    \setlength{\tabcolsep}{1pt}
    \begin{tabular}{cc|cc}
        \toprule
        \textbf{Image} & \textbf{Attention Map} & \textbf{Image} & \textbf{Attention Map}\\
        \midrule
        \multicolumn{2}{p{\llw\linewidth}|}{\centering  
        A \colorbox{lightgray}{\bf white} sink next to a toilet} & \multicolumn{2}{p{\llw\linewidth}}{\centering  \colorbox{lightgray}{\bf Plant} in window} \\
        \midrule
        \includegraphics[width=\lw\linewidth]{figures/attn-visualization/image27.png} & 
        \includegraphics[width=\lw\linewidth]{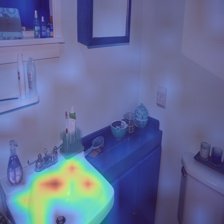} & 
        \includegraphics[width=\lw\linewidth]{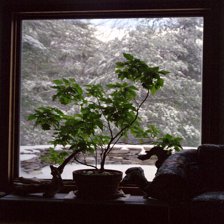} & 
        \includegraphics[width=\lw\linewidth]{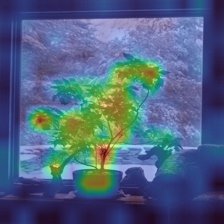}\\
        \midrule        \multicolumn{2}{p{\llw\linewidth}|}{\centering  A \colorbox{lightgray}{\bf cat} is napping with \\ a pile of sneakers} & \multicolumn{2}{p{\llw\linewidth}}{\centering Look up and see these beautiful \colorbox{lightgray}{\bf palms} in the breeze} \\
        \midrule
        \includegraphics[width=\lw\linewidth]{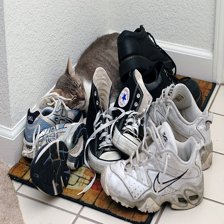} & 
        \includegraphics[width=\lw\linewidth]{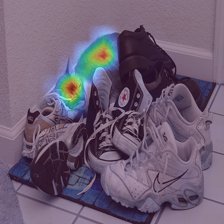} &
        \includegraphics[width=\lw\linewidth]{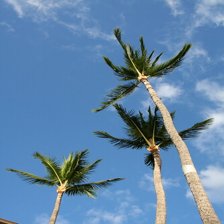} & 
        \includegraphics[width=\lw\linewidth]{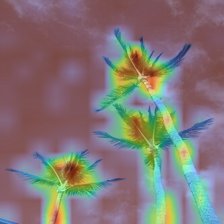} \\
        \midrule
        \multicolumn{2}{p{\llw\linewidth}|}{\centering  Person \colorbox{lightgray}{\bf walking} past \\ the grand temple} & \multicolumn{2}{p{\llw\linewidth}}{\centering 
        View down the main \colorbox{lightgray}{\bf street} from atop roman structure} \\
        \midrule
        \includegraphics[width=\lw\linewidth]{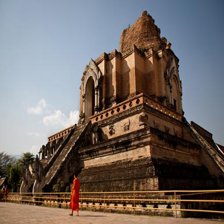} & 
        \includegraphics[width=\lw\linewidth]{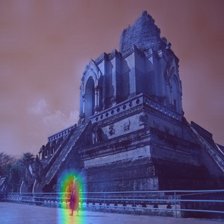} &
        \includegraphics[width=\lw\linewidth]{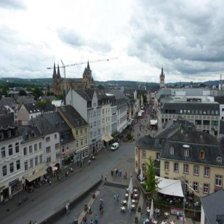} & 
        \includegraphics[width=\lw\linewidth]{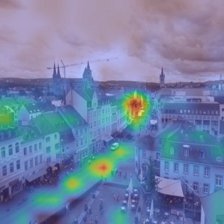} \\
        \bottomrule
        
    \end{tabular}
    \caption{Attention maps on predicting \colorbox{lightgray}{\bf masked} words through MoMo at stage 3. Heatmaps are obtained through Transformer-MM-Explainability~\cite{chefer2021generic}. MoMo is able to capture meaningful regions for MLM through cross-modality attentions.}
    \vspace{-10pt}
    \label{fig:momo-vis}
\end{figure}

\textbf{Simultaneous Training with Multiple Modalities.} For this ablation study, we investigate the effect of not doing simultaneous training as described in Sec.~\ref{sec:cmga} i.e. during stage 2, the model is trained only on the text dataset (image dataset is removed). And during stage 3, the model is trained only on multimodal datasets (text-only dataset is removed). Table \ref{tab:main-ablate} shows the performance difference between MoMo and those variations. At stage 2, performance on language tasks remains similar without simultaneous training, while its vision counterpart drops from 52.9\% to 9.5\%. Such drop indicates that learning text features can result in loss of image information if image data are not involved at that stage. Similarly, with a lower magnitude, results on text tasks degrade without simultaneous training at Stage 3. 

\textbf{Cross-Modality Gradient Accumulation.} The ``CMGA'' column in Table \ref{tab:main-ablate} shows that MoMo consistently achieves better performance on all tasks with Cross-modality gradient accumulation (Sec.~\ref{sec:cmga}).

\textbf{Shared Decoders.} As described in Sec.~\ref{fig:arch}, we adopt separate decoders for different modalities. In the ``Decoder'' column of Tab.~ \ref{tab:main-ablate}, we show the performance comparison if the decoder is, instead, shared at stage 2. Sharing the decoder results in a slight degradation in average performance.

\textbf{MLM with cross-modality knowledge.} As discussed in Sec.~\ref{fig:arch}, we use a cross-modal masking objective to enrich multimodal features. In Fig.~\ref{fig:momo-vis}, we visualize the attention map on images for MLM predictions. The results suggest that the model is able to capture the corresponding region in an image given a text representing that.

\section{Conclusion}
\label{sec:conclusion}
To summarize, we made three key contributions: i) We propose a compact and efficient transformer model that can be applied to vision-only, language-only and vision-language tasks without requiring modality-specific encoders; ii) We propose a new training pipeline that involves simultaneous learning from multiple modalities via a hierarchical stage-wise training; iii) By leveraging masked input reconstruction, contrastive and image-text matching objectives, MoMo is competitive with the state-of-the-art multimodal models that are both larger in size and are trained on much larger datasets. We also verify the scalability of MoMo through experiments with a larger transformer backbone. We present quantitative comparisons against similar sized baseline models demonstrating MoMo's effectiveness. We further present ablation studies to illustrate the impact of different design choices. In future, we plan to extend MoMo into larger models, include more pre-training data and incorporate extra modalities.

{\small
\bibliographystyle{ieee_fullname}
\bibliography{egbib}
}

\end{document}